\documentclass[letterpaper, 10 pt, conference]{ieeeconf}  

\IEEEoverridecommandlockouts

\usepackage{graphics} 
\graphicspath{{figures/}}
\usepackage{epsfig} 
\usepackage{mathptmx} 
\usepackage{times}
\usepackage{amsmath}
\usepackage{amssymb} 
\usepackage{cite}
\usepackage{bm}
\usepackage{url}
\usepackage[ruled,vlined,linesnumbered]{algorithm2e}
\usepackage{mathrsfs}
\let\labelindent\relax
\usepackage{enumitem}
\usepackage{subcaption}
\usepackage{booktabs}
\usepackage{multirow}

\newcommand{\vect}[1]{\boldsymbol{#1}}
\newcommand{\mat}[1]{\mathbf{#1}}

\title{\LARGE \bf
CSSDF-Net: Safe Motion Planning Based on Neural Implicit Representations of Configuration Space Distance Field

}

\author{Haohua Chen$^{1,*}$, Yixuan Zhou$^{1,*}$, Yifan Zhou$^{1}$ and Hesheng Wang$^{1,\dagger}$
\thanks{$^1$ Authors are with the IRMV Lab, Department of Automation, Shanghai Jiao Tong University, Shanghai 200240, China.} 
} 

\begin{document}
\maketitle
\thispagestyle{empty}
\pagestyle{empty}

\begin{abstract}
High-dimensional manipulator operation in unstructured environments requires a differentiable, scene-agnostic distance query mechanism to guide safe motion generation. Existing geometric collision checkers are typically non-differentiable, while workspace-based implicit distance models are hindered by the highly nonlinear workspace--configuration mapping and often suffer from poor convergence; moreover, self-collision and environment collision are commonly handled as separate constraints. We propose Configuration-Space Signed Distance Field-Net (CSSDF-Net), which learns a continuous signed distance field directly in configuration space to provide joint-space distance and gradient queries under a unified geometric notion of safety. To enable zero-shot generalization without environment-specific retraining, we introduce a spatial-hashing-based data generation pipeline that encodes robot-centric geometric priors and supports efficient retrieval of risk configurations for arbitrary obstacle point sets. The learned distance field is integrated into safety-constrained trajectory optimization and receding-horizon MPC, enabling both offline planning and online reactive avoidance. Experiments on a planar arm and a 7-DoF manipulator demonstrate stable gradients, effective collision avoidance in static and dynamic scenes, and practical inference latency for large-scale point-cloud queries, supporting deployment in previously unseen environments.
\end{abstract}

\section{INTRODUCTION}
As robots increasingly transition from controlled industrial settings to highly complex and unstructured real-world environments, the development of robust collision avoidance mechanisms has become increasingly essential. 

To generate feasible trajectories that simultaneously satisfy safety and smoothness constraints, nonlinear optimization-based backend trajectory optimization algorithms are typically employed. The core of these methods lies in the gradient-guided escaping from unsafe region. However, in practical, the accurate computation of distance gradients for complex shapes with compact geometric representations (e.g., convex hulls) poses significant challenges, particularly when handling dynamic obstacles that require complex mathematical derivations. 

Recently, implicit modeling strategies becomes a main solution to obtain differentiable, and high-fidelity representations of complex geometric boundaries. Specifically, distance proxy models can be constructed to represent the relationship between the robot and collision objects. However, due to the highly nonlinear mapping between the workspace and the configuration space (C-space), such methods are susceptible to local minima for manipulators. This has motivated the direct modeling of distance fields in C-space, thereby avoiding the inherent limitations of workspace-based methods caused by inverse-kinematics projections. However, existing approaches in C-Space still face two major challenges:

\textbf{Challenge 1: How to achieve a scenario-agnostic neural implicit representation without retraining across different environments?} A cascaded frameworks that map a pre-trained workspace SDF to C-space\cite{CDF} may complicates the pipeline by introducing additional pre-training stages, and exacerbates error accumulation across modules, ultimately degrading robustness and generalization.

\textbf{Challenge 2: How to unify self-collision and environment collision into a single geometric notion of safety?} Prior work often models these constraints separately---for example, assuming self-collision is implicitly avoided within manufacturer-defined joint limits, or training solely for self-collision. 

To solve these two challegens, we design Configuration-Space Signed Distance Field-Net (CSSDF-Net) , a deep neural network (DNN) to provide continuous distance/gradient query directly in C-space. Our core contributions are as follows:

\begin{itemize}
    \item CSSDF-Net is introduced to model a continuous, differentiable signed distance field directly in configuration space, providing a unified geometric notion of safety for both self-collision and environment collision.
    
    \item A spatial-hashing data generation pipeline is developed to train a scene-agnostic model, removing the need for environment-specific retraining and cascaded SDF mappings.
    
    \item The learned distance field and gradients are integrated into vision-guided motion planning and trajectory optimization, enabling real-time collision checking and safe motion generation in static and dynamic scenes.
\end{itemize}

\section{Related Works}
\label{related works}
Current researches on robot collision avoidance can be split into two main branches, free space approximation and distance field modeling. 
\subsection{Free Space Approximation}
Early studies, such as sampling-based planning algorithms, a series of methods developed from probabilistic roadmaps (PRM \cite{PRM}) and rapidly exploring random trees (RRT \cite{RRT}), construct a roadmap or a tree by randomly sampling collision-free configurations via a binary feasibility test and then search for a path; in essence, this amounts to an approximate construction of the topological structure of the free space. Similarly, this approximation paradigm has been widely adopted in mobile robotics, particularly in autonomous driving and unmanned aerial vehicles (UAVs). In these domains, the approach is often referred to as a safe flight corridor, where simple structures such as spheres and cubes are used in the task space\cite{online_quadrotor_traj,online_safe_traj,flying_on_point_clouds_online_traj} to construct a collection of collision-free regions as the feasible set, and this idea has been further extended to more general convex polytopes \cite{Planning_dynamically_feasible_traj_using_SFC,IRIS}. Although the above methods are difficult to apply directly to high-dimensional problems due to the “curse of dimensionality,” techniques such as mathematical mappings can be leveraged to transfer them to the configuration space\cite{certified_polyhedral_decomp_of_collision-free_cs,growing_convex_collision-free_regions_in_cs,super_fast_cs_convex_set_compute}.

\subsection{Distance Field Modeling}
Distance modeling methods can be categorized into two types: explicit and implicit. Explicit computation often employs geometric intersection algorithms such as GJK\cite{cameronEnhancingGJKComputing1997}. However, GJK is not mathematically differentiable and typically provides only a binary collision verdict and a distance value in FCL\cite{FCL}, PyBullet\cite{BulletPS}, lacking continuous information to guide downstream application. Consequently, implicit approaches that fit Euclidian Signed Distance Field (ESDF) with neural networks have emerged and have been successfully applied to scenarios such as autonomous driving and UAVs\cite{oleynikovaContinuoustimeTrajectoryOptimization2016,gaoGradientbasedOnlineSafe2017}. 

For high-dimensional problems, the nonlinear mapping between task space and configuration space makes task-space distance fields difficult to use. Recent studies suggest that modeling collisions directly in the robot configuration space\cite{vasilopoulosRAMPHierarchicalReactive2023} effectively overcomes inherent limitations of kinematics-based collision checking, such as singularities and local convergence issues.
For static environments or self-collision scenarios, existing methods primarily sample the robot joint space and employ machine-learning techniques to represent collision boundaries\cite{koptevRealtimeSelfcollisionAvoidance2021}. For dynamic environments, configuration-space distance field (CDF)\cite{liConfigurationSpaceDistance2024a} models the robot as a particle system where obstacles form explicit topological holes in the configuration space. The CDF approach is highly compatible and scalable, seamlessly integrating with modern geometric motion-planning frameworks such as Riemannian motion policies\cite{ratliffRiemannianMotionPolicies2018}, geometric fabrics\cite{vanwykGeometricFabricsGeneralizing2022}. 
\section{Methodology}
\label{method}

\subsection{Problem Formulation} 
Consider a robot with $m$ joints, of which $d$ are actively actuated. The remaining $m-d$ joints are passive or fixed. Since motion involves only the active joints, the robot configuration is uniquely defined by the vector $\vect{q}\in\mathbb{R}^{d}$. The corresponding topological space $\mathcal{Q}\subset \mathbb{R}^{d}$ is termed the configuration space.

Assume the robot comprises $k_r>m$ rigid links, each with $h>0$ attached collision bodies, denoted by $\mathcal{C}_i=\{c_i^1,\ldots,c_i^h\}$. A generalized self-collision distance function incorporating both link-to-link collisions and joint limits is defined as:
\begin{equation}\label{def:self-collision-distance}
d_{s}(\vect{q})=
\begin{cases}
\displaystyle\min_{\vect{q}_{c}\in\mathcal{Q}^{c}}\bigl\|\vect{q}-\vect{q}_{c}\bigr\|_{2}, & \vect{q}\in\mathcal{Q}^{s},\\[6pt]
-\displaystyle\min_{\vect{q}_{f}\in\mathcal{Q}^{s}}\bigl\|\vect{q}-\vect{q}_{f}\bigr\|_{2}, & \vect{q}\in\mathcal{Q}^{c}.
\end{cases}
\end{equation} 
where $\mathcal{Q}^{s}$ and $\mathcal{Q}^{c}$ denote the safe and self-collision configuration sets, respectively. 

 To establish a single geometric notion of safety, we treat pure self-collision as a special case of external collision where the obstacle point lies outside the reachable workspace. Suppose the workspace contains $g>0$ static or dynamic obstacles, denoted by $\mathcal{O}=\{O_1,\ldots,O_g\}$. For an arbitrary point $\vect{p}\in\mathbb{R}^{3}$ (either on an obstacle or virtual for self-collision), the signed distance function is given by:
\begin{equation}\label{def:out-collision-distance}
d_{c}(\vect{p},\vect{q})=
\begin{cases}
\displaystyle\min_{\vect{q}_{c}\in\mathcal{Q}_{c}^{p}}\|\vect{q}-\vect{q}_{c}\|_{2}, & \vect{q}\notin\mathcal{Q}_{c}^{p},\\[6pt]
-\displaystyle\min_{\vect{q}_{f}\notin\mathcal{Q}_{c}^{p}}\|\vect{q}-\vect{q}_{f}\|_{2}, & \vect{q}\in\mathcal{Q}_{c}^{p}.
\end{cases}
\end{equation}
where $\mathcal{Q}_{c}^{p}$ represents the set of configurations colliding with point $\vect{p}$. For self-collision, $\vect{p}$ is treated as a virtual point with $\mathcal{Q}_{c}^{p} \equiv \mathcal{Q}^{c}$.

Combining Eq.~\ref{def:self-collision-distance} and Eq.~\ref{def:out-collision-distance} yields the composite signed distance function for a single point $\vect{p}$:
\begin{equation}\label{def:cdf}
d(\vect{p},\vect{q})=
\begin{cases}
\max\bigl(d_{s}(\vect{q}),\,d_{c}(\vect{p},\vect{q})\bigr), & d_{s}(\vect{q})<0\;\land\;d_{c}(\vect{p},\vect{q})<0,\\[4pt]
\min\bigl(d_{s}(\vect{q}),\,d_{c}(\vect{p},\vect{q})\bigr), & \text{otherwise}.
\end{cases}
\end{equation}
The $\max$ operation ensures the most critical collision source dominates when both self-collision and external collision occur simultaneously, while $\min$ selects the nearest safety boundary otherwise.

Extending this to $k_{o}$ discrete obstacle points $\mathcal{C}=\{\vect{p}_{1},\ldots,\vect{p}_{k_{o}}\}$ in the workspace, the composite signed distance function is expressed as:
\begin{equation}
d(\mathcal{C},\vect{q})=
\begin{cases}
\displaystyle\max_{i=1,\ldots,k_{o}}d(\vect{p}_{i},\vect{q}), & \forall\,d(\vect{p}_{i},\vect{q})<0,\\[6pt]
\displaystyle\min_{i=1,\ldots,k_{o}}d(\vect{p}_{i},\vect{q}), & \text{otherwise}.
\end{cases}
\end{equation}

\begin{figure}[htb]
    \centering
    \includegraphics[width=1.0\linewidth]{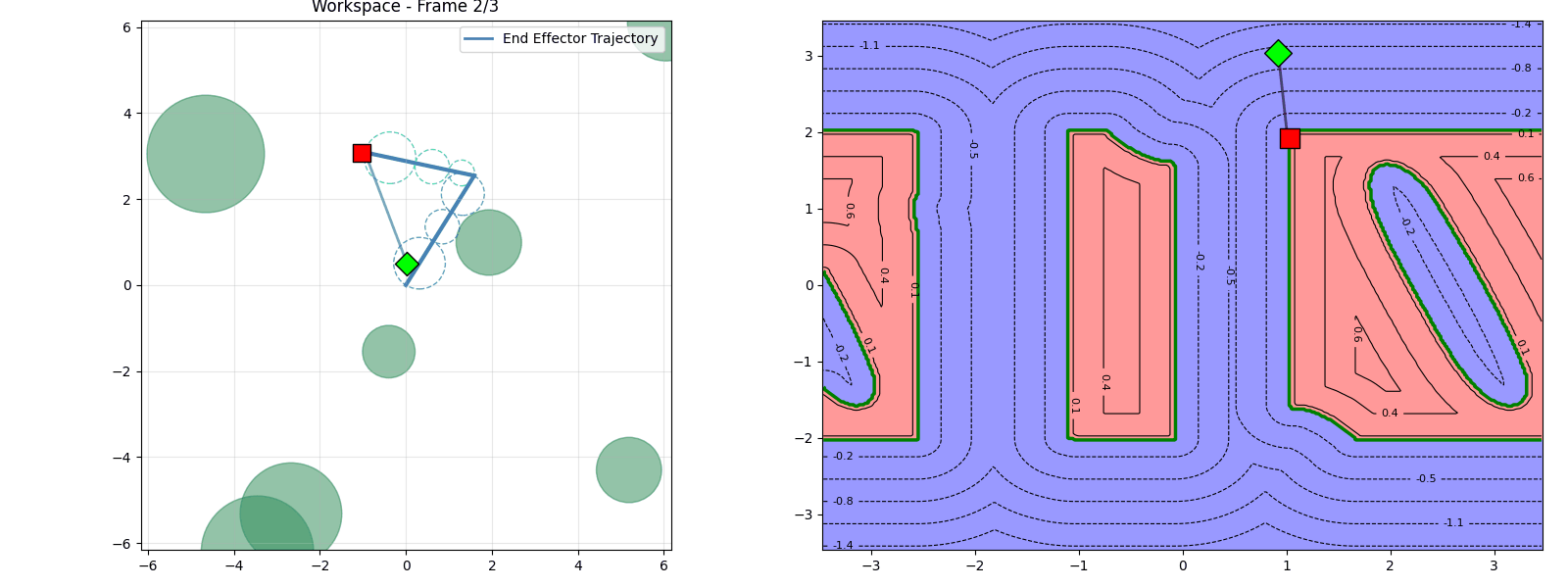}
    \caption{Gradient-guided escape in configuration space. Unlike Euclidean space methods requiring iterative projections through inverse kinematics (multi-step, curved paths), our C-space SDF enables direct single-step escape along the configuration gradient.}
    \label{fig:gradient_escape}
\end{figure} 

The signed distance function exhibits the following linear homogeneity properties:
\begin{equation}
\bigl\|\nabla_{\vect{q}} d(\vect{p},\vect{q})\bigr\|_{2}=1,
\vect{q}_{\mathrm{boundary}}=\vect{q}_{1}-d\bigl(\vect{p},\vect{q}_{1}\bigr)\nabla_{\vect{q}} d\bigl(\vect{p},\vect{q}_{1}\bigr),
\end{equation}
where the first property indicates unit Lipschitz continuity, and the second provides an explicit expression for reaching the collision boundary along the gradient direction.

Fig.~\ref{fig:gradient_escape} illustrates the key advantage of configuration-space representation: the initial robot configuration (green end-effector) inside an obstacle escapes to a safe region (red end-effector) along the shortest path in $\mathcal{Q}$, requiring only a single linear iteration. This contrasts with workspace-based methods that require multiple inverse-kinematics projections, resulting in curved, inefficient escape trajectories.
\subsection{CSSDF-Net Architecture}
\subsubsection{Network Structure}
CSSDF-Net takes concatenated input $\vect{x}=[\vect{q};\vect{p}]\in\mathbb{R}^{n+3}$, where $\vect{q}\in\mathbb{R}^{n}$ is the configuration vector and $\vect{p}\in\mathbb{R}^{3}$ is the obstacle point (or virtual point for self-collision). The network employs a 5-layer MLP backbone (216 neurons per layer) with identity residual connections, utilizing ReLU activations, batch normalization, positional encoding, and dropout ($p=0.2$). To balance the distribution of safe and collision data, inputs are min-max normalized to $[-1,1]$ using extended joint limits $[\min(-\pi,\vect{q}_{\min}),\max(\pi,\vect{q}_{\max})]$. The corresponding inverse transform of the gradient is applied via the chain rule during backpropagation to maintain optimization stability. The model is trained using the AdamW optimizer~\cite{AdamWOpt} with a ReduceLROnPlateau scheduler.

The network is supervised by a composite loss function $\mathcal{L} = 5\mathcal{L}_{\mathrm{dist}} + 0.1\mathcal{L}_{\mathrm{eikonal}} + 0.2\mathcal{L}_{\mathrm{dir}}$, comprising three terms:

\textbf{1) MSE Regression Loss} ($\mathcal{L}_{\mathrm{dist}}$) minimizes the error between the predicted $\hat{d}(\vect{x}_i)$ and ground-truth $d(\vect{x}_i)$ CSSDF values:
\begin{equation}
\mathcal{L}_{\mathrm{dist}}=\frac{1}{b}\sum_{i=0}^{b-1}\!\bigl(\hat{d}(\vect{x}_i)\!-\!d(\vect{x}_i)\bigr)^{2}.
\end{equation}

\textbf{2) Eikonal Regularization Loss} ($\mathcal{L}_{\mathrm{eikonal}}$) enforces the unit-gradient geometric property ($\lVert\nabla d\rVert\!=\!1$) of the SDF:
\begin{equation}
\mathcal{L}_{\mathrm{eikonal}}=\frac{1}{b}\sum_{i=0}^{b-1}\!\left(\lVert\nabla_{\vect{q}}d(\vect{x}_i)\rVert_{2}-1\right)^{2}.
\end{equation}

\textbf{3) Gradient Direction Consistency Loss} ($\mathcal{L}_{\mathrm{dir}}$) aligns the predicted and true gradient directions via cosine similarity to ensure reliable downstream motion planning:
\begin{equation}
\mathcal{L}_{\mathrm{dir}}=\frac{1}{b}\sum_{i=0}^{b-1}\!\left(1-
\frac{\nabla_{\vect{q}}\hat{d}(\vect{x}_i)^{\!\top}\nabla_{\vect{q}}d(\vect{x}_i)}
{\lVert\nabla_{\vect{q}}\hat{d}(\vect{x}_i)\rVert\,\lVert\nabla_{\vect{q}}d(\vect{x}_i)\rVert}\right)^{2}.
\end{equation}
\subsection{Dataset Construction}\label{dataset-construct} 
The dataset generation pipeline produces unified $(\vect{q},\vect{p})$ tuples for both self-collision and external collision scenarios.
\subsubsection{Self-Collision Dataset Construction} 
The configuration space $\mathcal{Q}\subset\mathbb{R}^{n}$ is uniformly sampled to generate a base configuration set ${Q}_{\mathrm{base}}\in\mathbb{R}^{N\times n}$, where $N$ is the sample size and $n$ denotes the degrees of freedom (DoF). An FCL-based collision checker $f_{\mathrm{col}}:\mathcal{Q}\rightarrow\{0,1\}$ labels each configuration as safe (0) or in-collision (1). To mitigate the class imbalance inherent in random sampling, an adaptive strategy is introduced: if the ratio of collision to free samples ($p_{\mathrm{col}}/p_{\mathrm{free}}$) exceeds a threshold $\tau=1.25$, neighborhood perturbation augmentation is applied.

To capture the complex, high-dimensional decision boundaries in the joint space, a boundary mining mechanism employing Hierarchical Navigable Small World (HNSW) graphs~\cite{HNSW} is utilized. HNSW provides logarithmic-complexity nearest-neighbor search suitable for high-dimensional boundary extraction. By maintaining separate search trees for collision and free samples, the nearest neighbor $\vect{q}_{i}^{\mathrm{nn}}$ of a configuration $\vect{q}_{i}$ in the opposing tree is queried. A bisection search between $\vect{q}_{i}$ and $\vect{q}_{i}^{\mathrm{nn}}$ then isolates the precise decision boundary point. Upon convergence, the new boundary sample $\vect{q}_{\mathrm{boundary}}$ is added to its respective tree.

The ground-truth distance and gradient are then obtained for any point $\vect{q}_i$:
\begin{equation}
\label{truth-gen}
    d(\vect{q}_i) = \lVert \vect{q}_i - \vect{q}_{\mathrm{boundary}}\rVert, \quad \nabla d_{\vect{q}} = \frac{1}{d}(\vect{q}_i - \vect{q}_{\mathrm{boundary}}).
\end{equation}

The final dataset comprises random, perturbed, and boundary samples, formulated as tuples $\bigl(Q,\textit{value},\textit{labels},\nabla\bigr)$. Here, $Q\in\mathbb{R}^{b\times(n+3)}$ denotes the concatenated configuration-obstacle vector, $\textit{value}\in\mathbb{R}^{b}$ is the signed distance, $\textit{labels}\in\{0,1\}^{b}$ is the binary collision label, and $\nabla\in\mathbb{R}^{b\times n}$ is the spatial gradient. To ensure that the self-collision notion could be visited correctly with no obstacles (outside the workspace), workspace coordinates are sampled from an extended space $\mathcal{W}_{\mathrm{ext}}$ rather than the strictly reachable space $\mathcal{W}_{\mathrm{reachable}}$. 

\subsubsection{External Collision Dataset Construction}
\label{sec:dataset_construction}
To achieve scene-agnostic neural implicit representation without environment-specific retraining, a configuration-to-workspace mapping is established that decouples robot-specific geometric priors from scene-dependent obstacle information. The core innovation lies in a spatial hashing mechanism that enables constant-time retrieval of risk configurations for arbitrary obstacle layouts.

\textbf{Spatial Hashing for Configuration-Voxel Association:}
\label{sec:spatial_hashing}
The robot geometry is approximated as sphere clusters via \texttt{SpherizedURDFGenerator}\footnote{\url{https://github.com/IRMV-Manipulation-Group/SpherizedURDFGenerator}}, yielding occupied region $\mathcal{O}(\boldsymbol{q})=\bigcup_{i=1}^{m}B(\boldsymbol{c}_i(\boldsymbol{q}),r_i(\boldsymbol{q}))$ for configuration $\boldsymbol{q}$. The workspace $\mathcal{W}\subset\mathbb{R}^3$ is discretized with resolution $\Delta x$, and a spatial hash function $h:\mathbb{R}^{3}\!\to\!\mathbb{Z}^{3}$ maps continuous positions to voxel indices:
$
    h(\boldsymbol{p})=\left\lfloor(\boldsymbol{p}-\boldsymbol{w}_{\min}) / \Delta x\right\rfloor.
$

For each collision sphere $B(\boldsymbol{c},r)$, the influenced voxel set $\mathcal{I}(B)$ is computed, establishing the grid-configuration mapping:
\begin{equation}
\label{eq:grid_config_mapping}
    \mathcal{M}(g)=\bigl\{\boldsymbol{q}\in\mathcal{Q}\mid \exists\,B\in\mathcal{O}(\boldsymbol{q}),\, g\in\mathcal{I}(B)\land B\cap g\neq\varnothing\bigr\}.
\end{equation}

\begin{algorithm}[t]
    \caption{Parallel Spatial Hashing Construction}
    \label{alg:spatial_hashing}
    \SetAlgoLined
    \KwData{Configurations $\mathcal{Q}$, kinematic mapping $f:\mathcal{C}\to 2^{\mathcal{B}}$}
    \KwResult{Mapping $\mathcal{M}:\mathcal{G}\to 2^{\mathcal{Q}}$}
    
    Initialize $\mathcal{M}\leftarrow\emptyset$\;
    \ForEach{$\boldsymbol{q}\in\mathcal{Q}$ \textbf{parallel}}{
        $\mathcal{B}\leftarrow f(\boldsymbol{q})$\;
        \ForEach{$B(\boldsymbol{c},r)\in\mathcal{B}$}{
            $\mathcal{I}(B)\leftarrow$ compute influence region\;
            \ForEach{$g\in\mathcal{I}(B)$}{
                \If{$B\cap g\neq\varnothing$}{
                    $\mathcal{M}(g)\leftarrow\mathcal{M}(g)\cup\{\boldsymbol{q}\}$
                }
            }
        }
    }
    \Return $\mathcal{M}$\;
\end{algorithm}

Algorithm~\ref{alg:spatial_hashing} constructs $\mathcal{M}$ offline via parallel computation. This mapping is computed \textit{once} during training and enables constant-time retrieval of all configurations potentially occupying any voxel $g$ at data generation

\textbf{Zero-Shot capabilities for Novel Environments.} 
Crucially, $\mathcal{M}$ encodes robot-centric geometric priors that are independent of specific obstacle placements. The spatial hashing mechanism discretizes the workspace into voxels solely to determine the key-problem: how to obtain the zero-level-set of obstacle point $p$ in configuration space. This problem was treated as a nonliner-projection problem in \cite{CDF} using a pre-trained ESDF model. For data generation, given point $p$, the voxel $g$ is found and queried in $\mathcal{M}$ to obtain $\mathcal{Q}_{c}^{p}$ in \eqref{def:out-collision-distance}. Similar to \eqref{truth-gen}, ground truth for external collision is produced. 

Notice that, although the input data is discretized, the trained MLP learns a continuous representation of this space, with its intrinsic function approximation capability providing smooth interpolation across the entire workspace. When deployed in novel environments, incoming obstacle point clouds are simply transformed into the robot-centric coordinate frame and fed directly as continuous coordinates $\vect{p} \in \mathbb{R}^3$ to the network. No voxel discretization is required at inference time---the network naturally generalizes to arbitrary obstacle positions within its learned spatial bounds, effectively activating the corresponding geometric memories encoded during training. For dynamic scenes, updated sensor observations are transformed and queried at each timestep without model retraining. 
\subsection{Motion Planning and Control Integrating CSSDF-Net}

\subsubsection{Safe MPC for Dynamic Environments}
To demonstrate the capability of handling dynamic environments, the differentiable distance field learned by CSSDF-Net is integrated into a Model Predictive Control (MPC) framework. By formulating a safety-constrained receding horizon optimization problem, this approach ensures that the system can reactively stay within the safe configuration space while satisfying kinematic limits during trajectory tracking. 

Consider an $n$-DOF manipulator system, whose discrete-time state function can be expressed as a single-integrator model: 
\(\vect{q}_{k+1} = \mat{A}\vect{q}_k + \mat{B}\vect{u}_k,\)
where $\vect{q}_k \in \mathbb{R}^n$ is the joint configuration at time step $k$, $\vect{u}_k \in \mathbb{R}^n$ is the control input (joint velocity), $\mat{A} = \mat{I}_n$ is the state transition matrix, $\mat{B} = \Delta t \cdot \mat{I}_n$ is the control input matrix, and $\Delta t$ is the sampling period. To maintain the safety margin $\phi(\vect{q}_k) \geq \gamma$, the following linearized inequality constraint must be satisfied:
\( \nabla\phi(\vect{q}_k)^\top (\vect{q}_{k+1} - \vect{q}_k) \geq \Delta t(\gamma - \phi(\vect{q}_k)).\) 
Therefore, we design a safe MPC controller to solve the following constrained optimization problem over the prediction horizon $H$: 
\begin{equation}
\begin{aligned}
\min_{\vect{u}} \quad & \sum_{k=0}^{H-1} \left( \| \vect{q}_k - \hat{\vect{q}}_{k} \|^2_{\mat{Q}} + \| \vect{u}_k \|^2_{\mat{R}} \right), \\
\text{s.t.} \quad & \vect{q}_{k+1} = \mat{A}\vect{q}_k + \mat{B}\vect{u}_k, \quad k = 0,\dots,H-1, \\
& \vect{q}_{\min} \leq \vect{q}_k \leq \vect{q}_{\max}, \quad k = 1,\dots,H, \\
& \vect{u}_{\min} \leq \vect{u}_k \leq \vect{u}_{\max}, \quad k = 0,\dots,H-1, \\
& \nabla\phi(\vect{q}_k)^\top (\vect{q}_{k+1} - \vect{q}_k) \geq (\gamma - \phi(\vect{q}_k))\Delta t, \quad k = 1,\dots,H-1,
\end{aligned}
\label{eq:mpc_problem}
\end{equation}
Here, $\hat{\vect{q}}_k$ denotes the reference point on the target trajectory, $\mat{Q}$ and $\mat{R}$ are positive definite weight matrices, and $\gamma > 0$ represents the preset safety margin threshold. We utilize an efficient Quadratic Programming (QP) solver, such as OSQP, to solve the optimization problem in \eqref{eq:mpc_problem} in real time.

\subsubsection{Safety-Constrained Trajectory Optimization for Offline Planning}\label{visual_safe_planning}
To evaluate the offline processing performance, we formulate a safety-constrained trajectory optimization framework based on CSSDF-Net. The core of this framework lies in the seamless integration of the learned distance field and its gradient information into the backend optimization. This enables efficient offline generation of safe trajectories by rigorously evaluating safety constraints, encompassing both external and self-collisions. The objective function for the spatiotemporal trajectory optimization is formulated as:
\begin{equation}
    \label{chap4-planing}
    \begin{aligned}
        \min_{\vect{q},\,T_i}\;&\sum_{i=0}^{n-1}\bigg(\lambda_s\dfrac{T_i}{3}\operatorname{tr}\left(\vect{m}_j^{T}\mat{K}_2\vect{m}_j\right)
        +\lambda_tT_i \\
        &\quad +\lambda_p\lVert T_i-T_{i-1}\rVert^{2} + \lambda_q \lVert \vect{q}_{i+1} - \vect{q}_i\rVert\bigg) + \lambda_{\phi}\mat{M}_{s}(\vect{q}),\\
        \text{s.t.}\;&\vect{q}_i\in[\vect{q}_{\min},\vect{q}_{\max}],\;
        \vect{m}_i\in[\ddot{\vect{q}}_{\min},\ddot{\vect{q}}_{\max}],\\
        &\dfrac{\vect{q}_{i+1}-\vect{q}_i}{T_i}-\dfrac{T_i}{3}\vect{m}_i-\dfrac{T_i}{6}\vect{m}_{i+1}
        \in[\dot{\vect{q}}_{\min},\dot{\vect{q}}_{\max}],\\
        &\dot{\vect{q}}_0 = \vect{v}_s, \dot{\vect{q}}_n=\vect{v}_e, T_i>0
    \end{aligned}
\end{equation}
where $\mat{K}_2=\begin{bmatrix}1&1\\0&1\end{bmatrix}$ is the weight matrix, $\vect{m}_j=[\vect{m}_i^{T},\vect{m}_{i+1}^{T}]^{T}$ represents the concatenated vector of accelerations at adjacent control points, $\vect{q} = [\vect{q}_0,\dots,\vect{q}_N]^T$ is the sequence of spline control points, and $\mat{T} = [T_0,\dots,T_N]$ is the vector of time allocation parameters. 

In this framework, the safety penalty term $\mat{M}_s$ utilizes a continuously differentiable piecewise exponential function:
\begin{equation}
\mat{M}_{s}(\vect{q}) = \sum_{k=0}^{n-1} 
\begin{cases} 
\exp\left( -\dfrac{\phi(\vect{q}_k)}{d_0} \right) - 1, & \phi(\vect{q}_k) < 0 \\
\exp\left( -\alpha (\phi(\vect{q}_k) + d_0) \right), & \phi(\vect{q}_k) \geq 0 
\end{cases}
\label{eq:safety_penalty}
\end{equation}
where $\phi(\cdot)$ denotes the configuration space signed distance queried directly from CSSDF-Net.

It is worth emphasizing that because CSSDF-Net provides an accurate and continuous characterization of the safety boundary, this offline method can aggressively minimize the trajectory length by incorporating the term $\lVert \vect{q}_{i+1} - \vect{q}_i\rVert$ into the objective function while strictly guaranteeing safety. This capability distinguishes our approach from classic optimization frameworks that often struggle to balance path optimality with rigorous collision avoidance.
\section{Experiments and Results}
\label{experiments}
\subsection{2D Experiments}
\begin{figure}[htb]
    \centering
    \begin{subfigure}[b]{1.0\linewidth}
        \includegraphics[width=\linewidth]{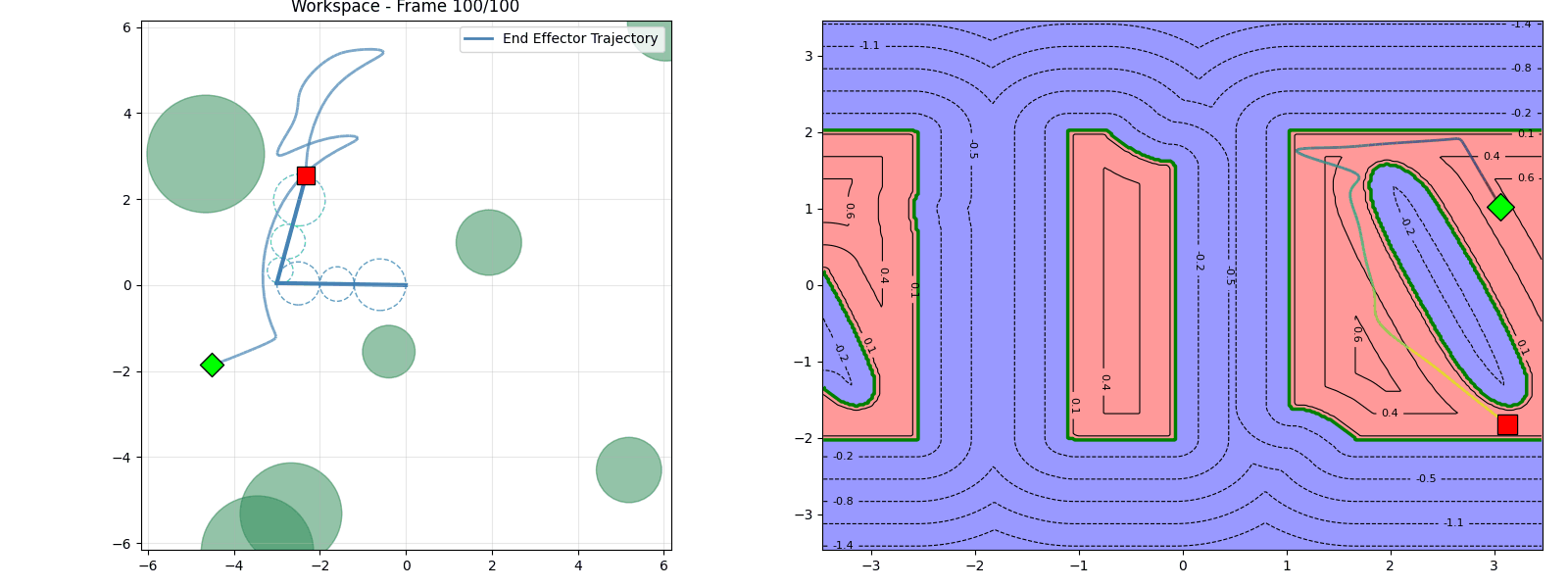}
        \caption{Sampling-based motion planning results}
        \label{sampling_results}
    \end{subfigure}
    \hfill
    \begin{subfigure}[b]{1.0\linewidth}
        \includegraphics[width=\linewidth]{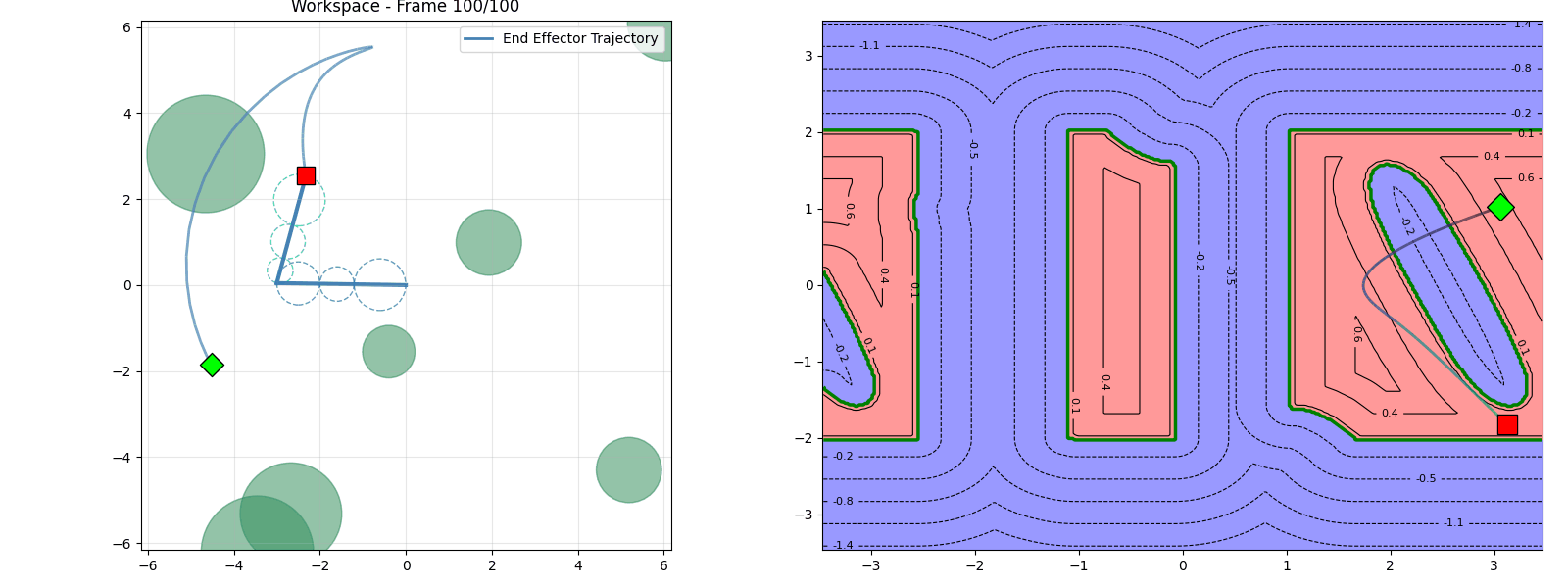}
        \caption{Optimization results without safety constraints}
        \label{direct_optmization}
    \end{subfigure}
    \hfill
    \begin{subfigure}[b]{1.0\linewidth}
        \includegraphics[width=\linewidth]{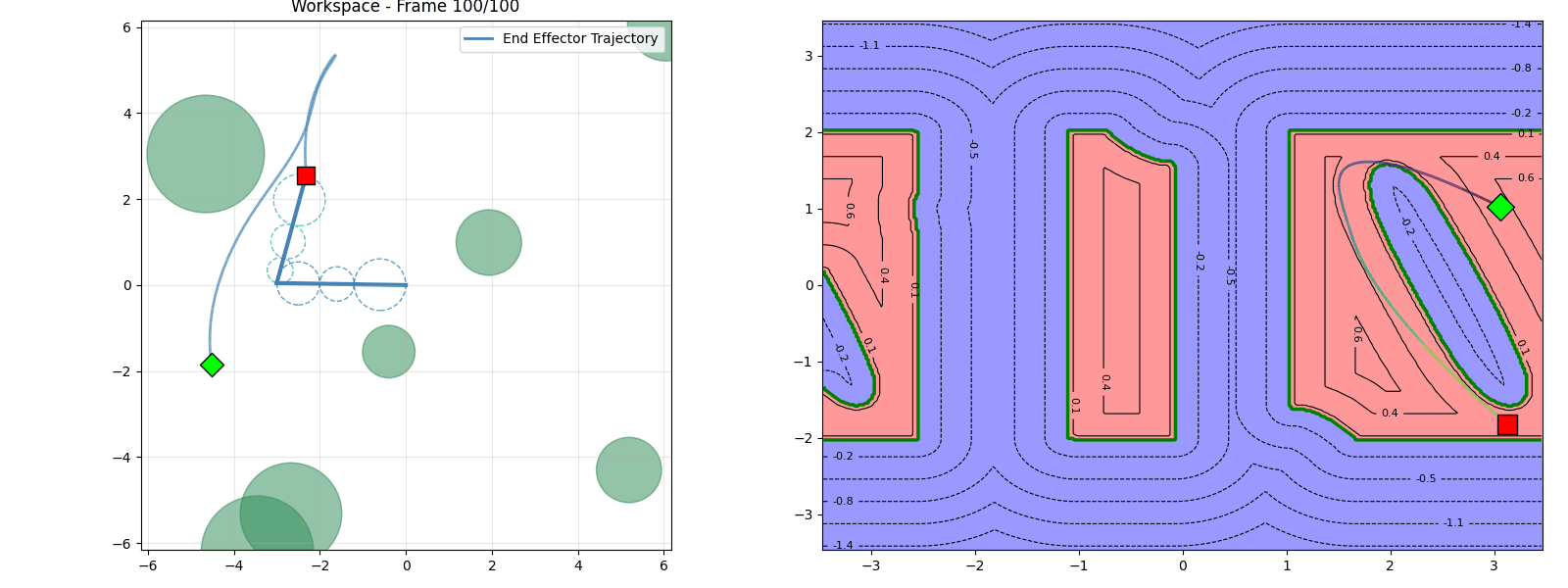}
        \caption{Optimization results with safety constraints}
        \label{safe_optmization}
    \end{subfigure}
    \caption{Comparative analysis of trajectory optimization with safety constraints.}
    \label{fig:optimization}
\end{figure}
To provide an intuitive validation of CSSDF-Net and its gradient-guided safety constraints, we first conduct experiments with a 2-DoF planar manipulator. Its configuration space is $\mathcal{Q}\subset\mathbb{R}^2$ with joint angles $[q_1,q_2]^\top\in[-\pi,\pi]^2$, and the workspace $\mathcal{W}\subset\mathbb{R}^2$ is a $12\times12~\text{m}^2$ region containing circular obstacles.
\subsubsection{Static Environment}
As shown in Fig.~\ref{fig:optimization}, sampling-based planners generate collision-free solutions but often produce jagged trajectories (Fig.~\ref{sampling_results}). In contrast, unconstrained trajectory optimization yields smooth paths but can penetrate obstacles due to the lack of a safety term (Fig.~\ref{direct_optmization}). By integrating CSSDF-Net queries $\phi(\vect{q})$ and $\nabla\phi(\vect{q})$ into the optimization, the resulting trajectories are both smooth and collision-free (Fig.~\ref{safe_optmization}), indicating that the learned C-space SDF provides usable gradients for escaping unsafe regions and converging to safe optima.

Table~\ref{tab:planar_planning} reports quantitative results in terms of Collision Risk (CR), Trajectory Length (TL), and Average Time (AT). The CSSDF-Net-guided safe optimization achieves $0\%$ CR with shorter trajectories than pure sampling, at the cost of higher computation time (130.48 ms). This trade-off reflects the additional safety evaluations while supporting our goal of enabling safety-critical, gradient-based trajectory optimization.
\begin{table}[h]
\centering
\caption{Comparative performance of trajectory planning for the planar manipulator}
\label{tab:planar_planning}
\begin{tabular}{lcccc}
\toprule
\multirow{2}{*}{Methods} & \multicolumn{3}{c}{Metrics} \\   
\cmidrule{2-4}
 & CR(\%)$\downarrow$ & TL$\downarrow$ & AT (ms)$\downarrow$ \\
\midrule
Sampling & 0. & 8.359 & \textbf{1.702} \\
Sampling + Non-safe optimization & 6.100 &  \textbf{5.793} & 3.216 \\
Sampling + Safe optimization & \textbf{0.} & 5.928 & 130.48 \\
\bottomrule
\end{tabular}
\end{table}
\subsubsection{Dynamic Environment}
We further evaluate the method in a dynamic environment with a $0.3~\text{m}$ rectangular obstacle moving leftward at a constant speed. At each control cycle, CSSDF-Net provides $\phi(\vect{q})$ and $\nabla\phi(\vect{q})$ for the observed obstacle points, which are used in the receding-horizon controller (Eq.~\eqref{eq:mpc_problem}) to enforce the linearized safety constraint.

\begin{figure}[htb]
    \centering
    \begin{subfigure}[b]{1.0\linewidth}
        \includegraphics[width=\linewidth]{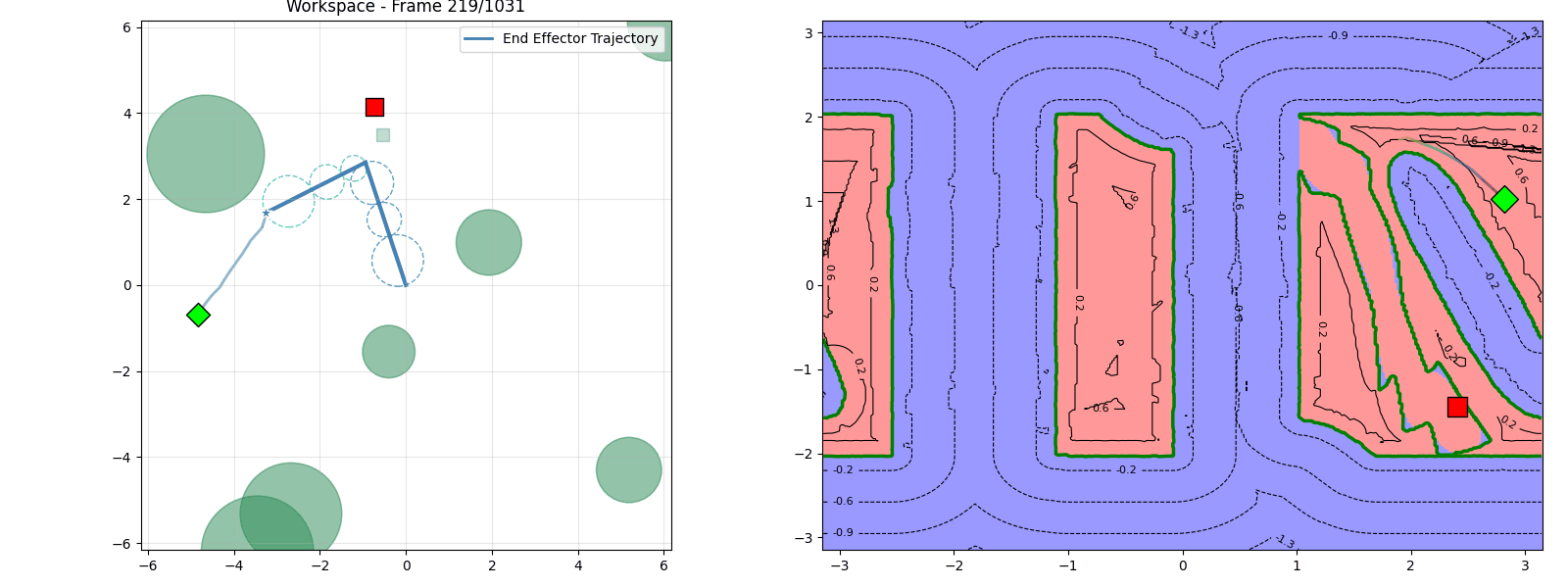}
        \caption{Obstacle enters the affected region}
        \label{first_obstacle}
    \end{subfigure}
    \hfill
    \begin{subfigure}[b]{1.0\linewidth}
        \includegraphics[width=\linewidth]{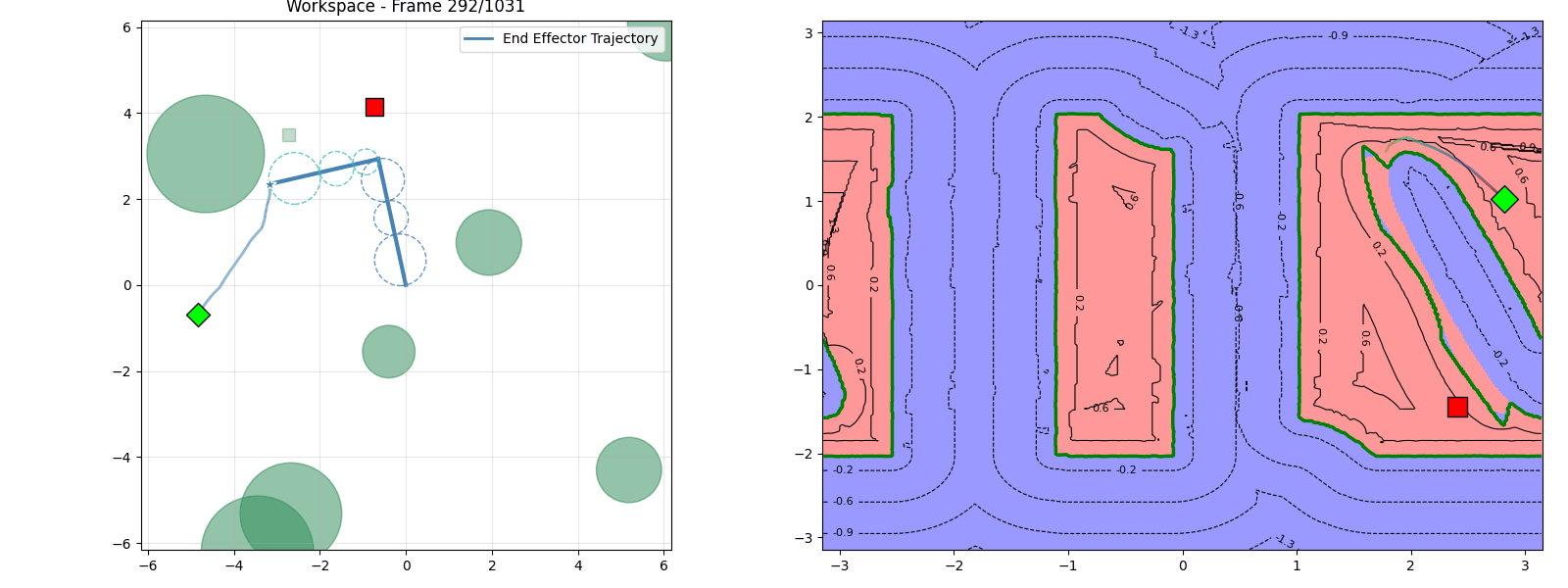}
        \caption{Obstacle leaves the affected region}
        \label{out_obstacle}
    \end{subfigure}
    \hfill
    \begin{subfigure}[b]{1.0\linewidth}
        \includegraphics[width=\linewidth]{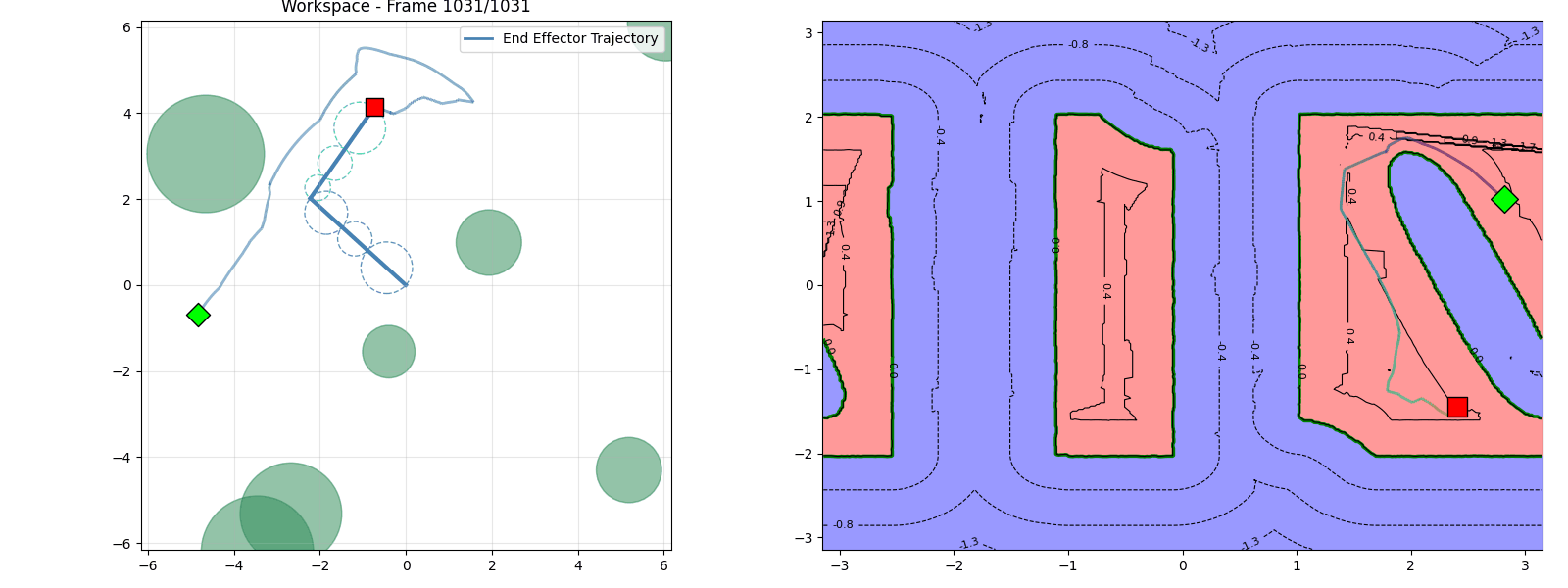}
        \caption{End-effector reaches the target point}
        \label{goal_reached}
    \end{subfigure}
    \caption{Evolution of the dynamic obstacle avoidance process.}
    \label{fig:dynamic_avoidance}
\end{figure}

Figure~\ref{fig:dynamic_avoidance} illustrates the avoidance process. When the obstacle enters the affected region (Fig.~\ref{first_obstacle}), the controller promptly generates a safe deviation. After the obstacle leaves (Fig.~\ref{out_obstacle}), the manipulator smoothly returns to its nominal motion and reaches the goal (Fig.~\ref{goal_reached}), demonstrating that the learned distance gradients remain stable for online reactive control.
\begin{table*}[htb]
\centering
\small 
\setlength{\tabcolsep}{4.5pt} 
\caption{Comprehensive Ablation Results on CSSDF-Net Components}
\label{tab:ablation-comprehensive}
\begin{tabular}{l ccc ccc}
\toprule
\multirow{2}{*}{Configuration} & \multicolumn{2}{c}{Primary Metrics} & \multicolumn{4}{c}{Secondary Specific Metrics} \\
\cmidrule(lr){2-3} \cmidrule(lr){4-7}
 & MAE (rad) $\downarrow$ & Grad. Sim. $\uparrow$ & FPR (\%) $\downarrow$ & Latency ($\mu$s) $\downarrow$ & BSR (\%) $\uparrow$ & Class Ratio (\%) \\
\midrule
\multicolumn{7}{c}{\textbf{Data Construction Strategy}} \\
\midrule
Uniform Sampling & 0.1256 & 0.8536 & -- & -- & 0.0000 & 86.9000 \\
Class-Balanced Sampling & 0.1029 & 0.8698 & -- & -- & 0.0117 & \textbf{50.8505} \\
Complete Strategy & \textbf{0.0754} & \textbf{0.8762} & -- & -- & \textbf{35.5656} & 51.7351 \\

\midrule
\multicolumn{7}{c}{\textbf{Loss Function}} \\
\midrule
Only Distance Loss & 0.1417  &  0.6645  & 15.4535 & -- & -- & -- \\
Distance + Magnitude Loss & 0.1369  & 0.6654  & 16.1909 & -- & -- & -- \\
Distance + Direction Loss & 0.1231  & 0.7517 & 11.7244 & -- & -- & -- \\
Complete Loss & \textbf{0.0754} & \textbf{0.8536} & \textbf{8.4167} & -- & -- & -- \\
\bottomrule
\multicolumn{7}{l}{\scriptsize \textit{Note:} MAE = Distance Mean Absolute Error; Grad. Sim. = Gradient Direction Similarity;} \\
\multicolumn{7}{l}{\scriptsize FPR = Boundary False Positive Rate; BSR = Boundary Sample Ratio.}
\end{tabular}
\end{table*}
\subsection{Ablation Studies}
Ablation experiments were performed on an NVIDIA RTX 3060 GPU using $1\times10^4$ samples with an 8:2 train--validation split. To evaluate reliability near the decision boundary, we define the False Positive Rate (FPR) as:
\begin{equation}
\mathrm{FPR} = \frac{\sum_{i=1}^{N_B}\mathbb{I}(\hat{d}(\vect{x}_i)<0 \cap d(\vect{x}_i)>0)}{N_B} \times 100\%
\end{equation}
where $N_B$ denotes the number of samples within 0.05 rad of the collision boundary and $\mathbb{I}(\cdot)$ is the indicator function.
\subsubsection{Dataset Construction Strategy}
The complete strategy increases the Boundary Sample Ratio (BSR) to 35.5656\%, whereas uniform sampling yields a 0\% BSR, indicating insufficient coverage of boundary regions. This improvement is accompanied by the lowest MAE and a near-balanced class ratio (51.7351\%), supporting the effectiveness of boundary mining for learning accurate C-space safety boundaries.
\subsubsection{Loss Functions}
Training with only distance loss yields low gradient similarity and a high FPR (15.45\%), which is undesirable for gradient-based planning and MPC linearization. Adding the direction loss improves gradient similarity and reduces FPR, while the eikonal (magnitude) term further stabilizes gradient behavior. Overall, the complete composite loss achieves the lowest MAE and reduces FPR to 8.42\%, providing more reliable gradients for downstream optimization.
\subsection{Real-World Manipulator Experiments}
To validate scalability in higher-dimensional configuration spaces, we conducted experiments on a 7-DoF manipulator (Fig.~\ref{fig:7dof_dynamic_avoidance}). The workspace includes static obstacles and human-driven dynamic obstacles, evaluating both trajectory optimization and online avoidance.
\subsubsection{Real-Time Inference Performance}
Table~\ref{tab:7d_inference} reports inference latency for distance and distance+gradient queries under different point-cloud sizes. GPU execution achieves millisecond-level latency for $10^2$--$10^4$ points (e.g., 4.302 ms at $10^4$ points for distance+gradient), enabling high-frequency replanning. CPU inference remains practical up to $10^3$ points (5.846 ms for distance+gradient), supporting real-time deployment under moderate sensing loads.
\begin{table}[h]
\centering
\caption{System inference performance for the 7-DoF manipulator}
\label{tab:7d_inference}
\resizebox{\columnwidth}{!}{%
\begin{tabular}{lcccccc}
\toprule
\multirow{2}{*}{Computing Mode} & \multicolumn{6}{c}{Query Scale (Number of Points)} \\
\cmidrule{2-7}
 & $10^0$ & $10^1$ & $10^2$ & $10^3$ & $10^4$ & $10^5$ \\
\midrule
Dis (CPU) & 0.302 & 0.367 & 0.633 & 2.431 & 28.721 & 286.053 \\
Dis+Grad (CPU) & 0.704 & 0.866 & 1.576 & 5.846 & 58.785 &  578.978 \\
Dis (GPU) & 1.185 & 1.102 & 1.107 & 1.174 & 2.118 & 22.193 \\
Dis+Grad (GPU) & 2.676 & 2.551 & 2.549 & 2.739 & 4.302 & 43.233 \\
\bottomrule
\end{tabular}%
} 
\end{table}
\subsubsection{Trajectory Optimization in Static Environments}
\begin{table}[t]
\centering
\caption{Trajectory planning performance for the 7-DoF manipulator}
\label{tab:7dof_planning}
\begin{tabular}{lcccc}
\toprule
\multirow{2}{*}{Methods} & \multicolumn{3}{c}{Metrics} \\   
\cmidrule{2-4}
 & CR(\%)$\downarrow$ & TL$\downarrow$ & AT(ms)$\downarrow$ \\
\midrule
Sampling & \textbf{0.274} & 34.279 & \textbf{8.846} \\
Sampling+Non-safe Optimization & 15.066 &  \textbf{ 17.898} & 31.738 \\
Sampling+Safe Optimization & 0.561 & 18.054 & 171.117 \\
\bottomrule
\end{tabular}
\end{table}

CSSDF-Net-guided safe optimization achieves low collision risk (0.561\%) and significantly shorter trajectories (18.054 rad) compared to pure sampling (34.279 rad) (Table~\ref{tab:7dof_planning}). Although safety constraints increase the computation time, the 171.117 ms latency is practical for offline planning.

\begin{figure*}[htb]
    \centering
    \begin{subfigure}[b]{0.32\linewidth}
        \includegraphics[width=\linewidth]{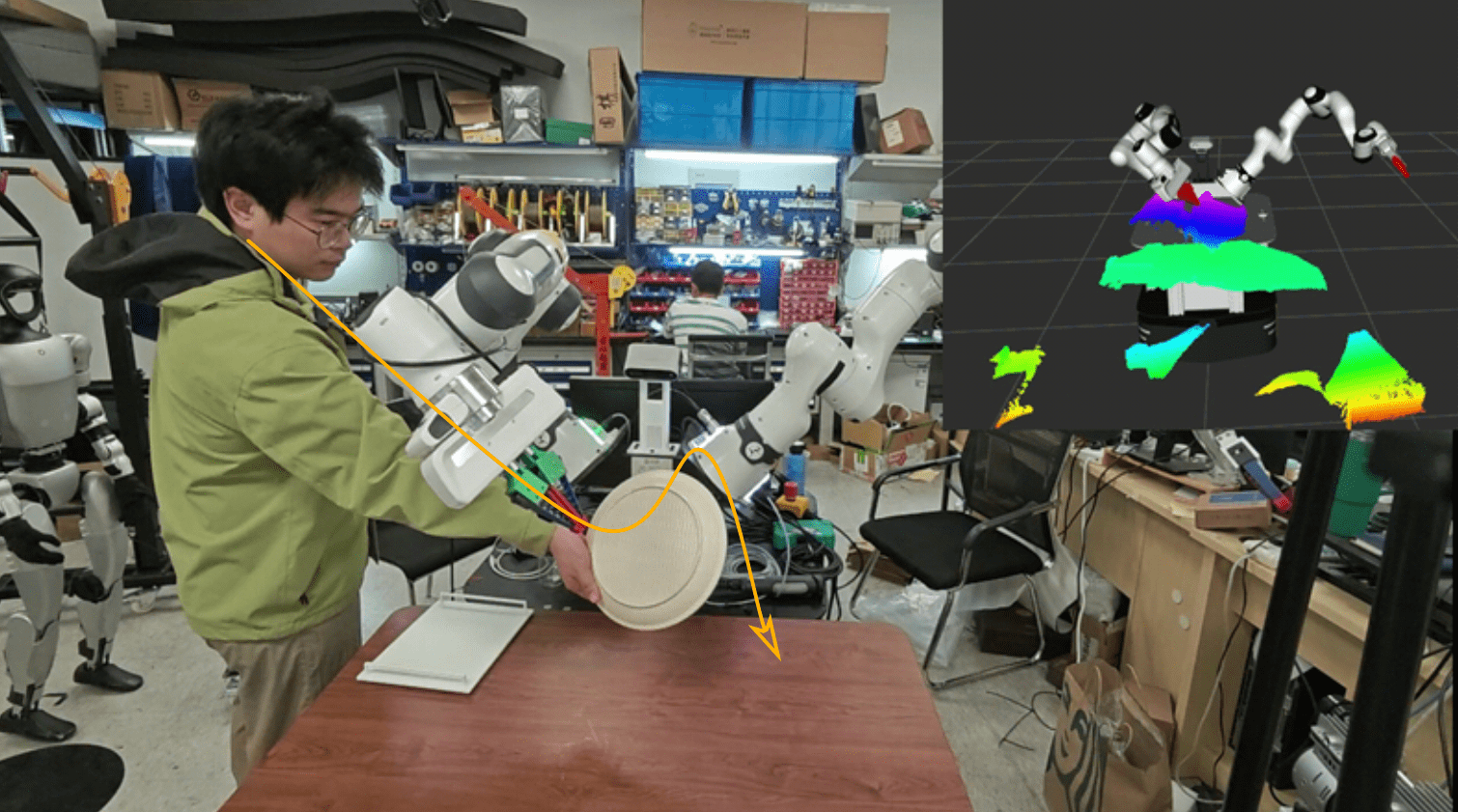}
        \caption{Obstacle enters affected region}
        \label{7dof_first_obstacle}
    \end{subfigure}
    \begin{subfigure}[b]{0.32\linewidth}
        \includegraphics[width=\linewidth]{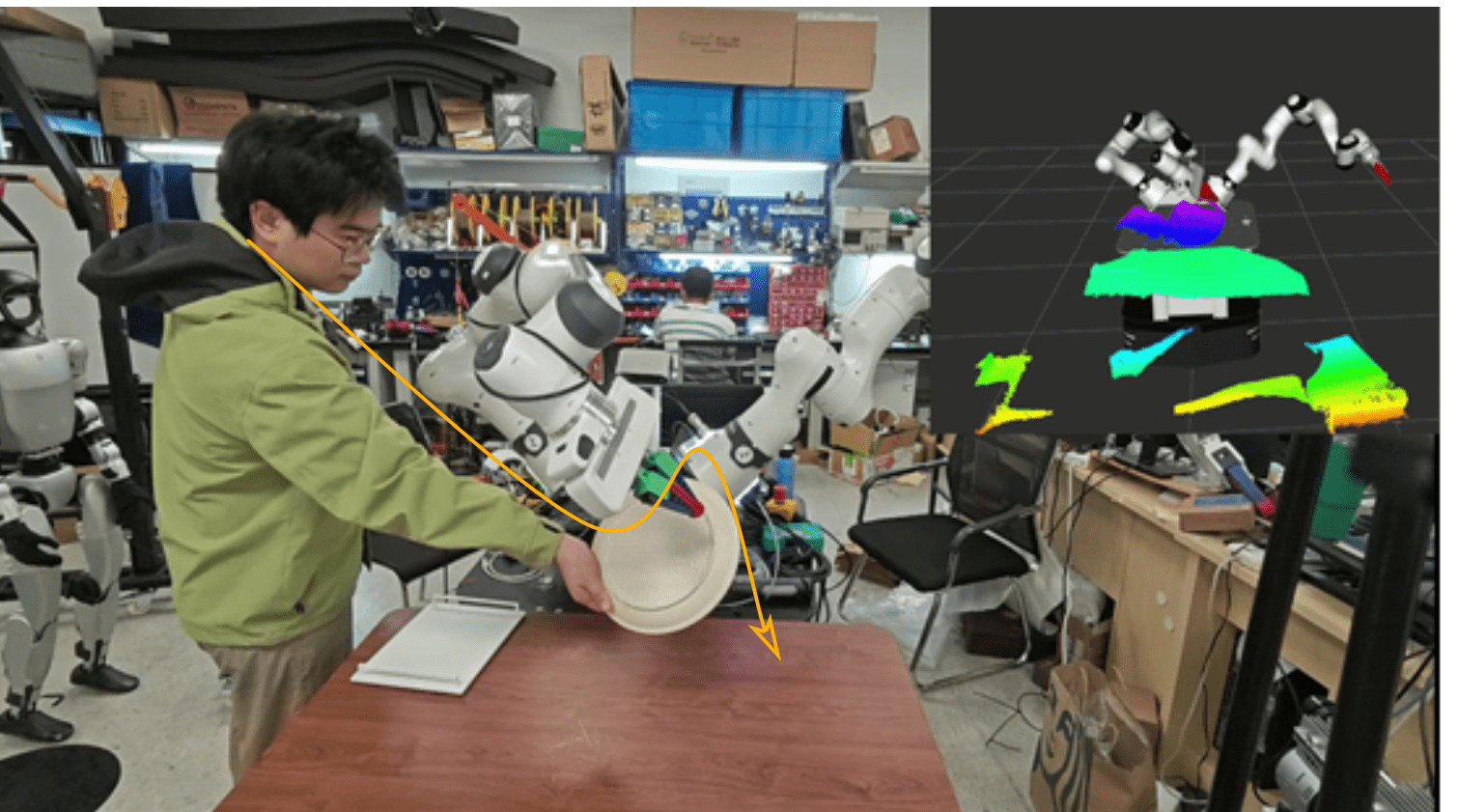}
        \caption{Obstacle leaves affected region}
        \label{7dof_out_obstacle}
    \end{subfigure}
    \begin{subfigure}[b]{0.32\linewidth}
        \includegraphics[width=\linewidth]{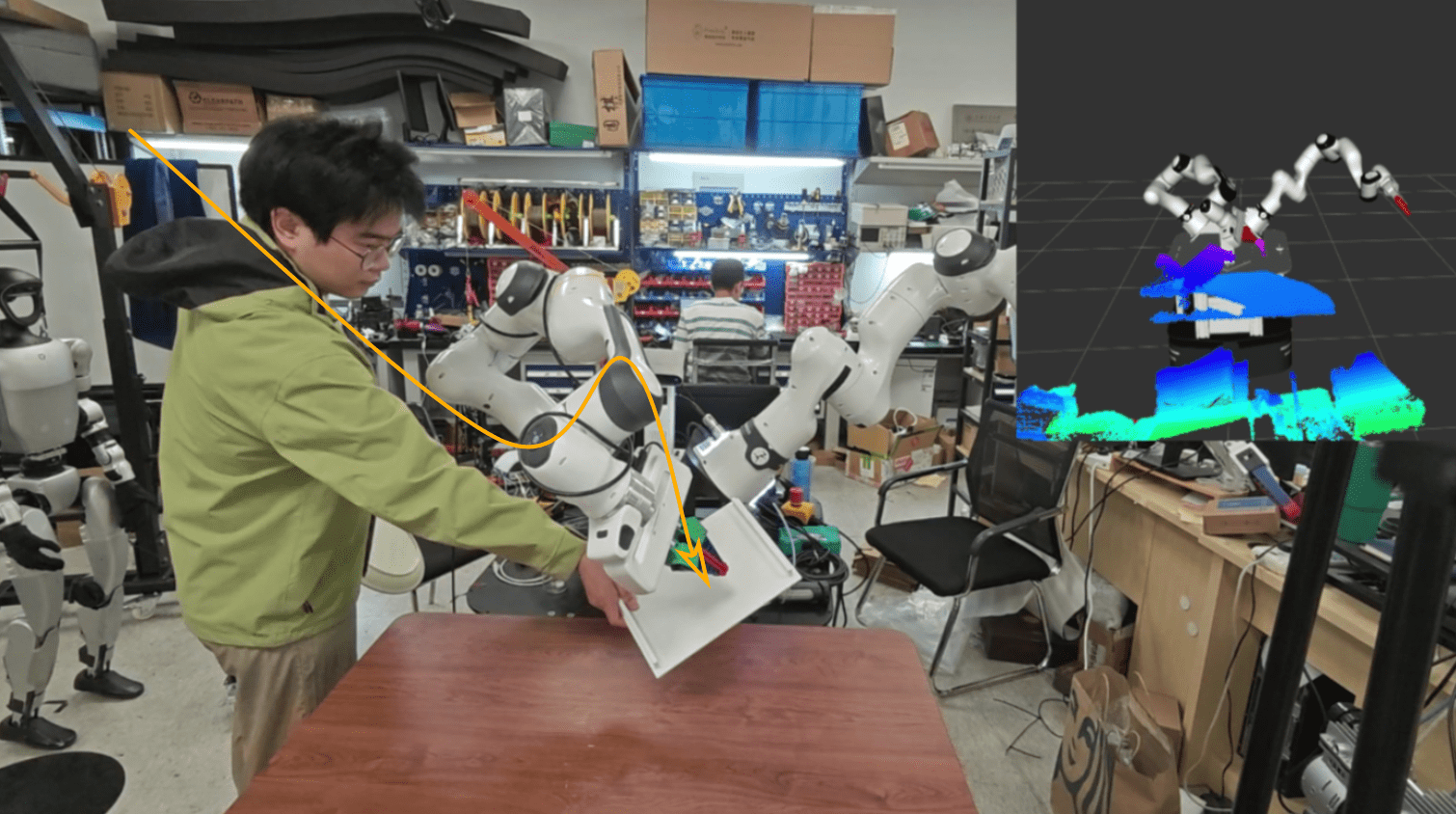}
        \caption{Manipulator reaches the goal}
        \label{7dof_goal_reached}
    \end{subfigure}
    \caption{Dynamic obstacle avoidance sequence for the 7-DoF manipulator.}
    \label{fig:7dof_dynamic_avoidance}
\end{figure*}
\subsubsection{Experiments in Dynamic Environments}
We integrated CSSDF-Net into the MPC controller to evaluate real-time avoidance using dynamic point clouds updated at 100 Hz. Table~\ref{tab:7dof_mpc_online} shows a trade-off governed by the receding horizon length ($H$): a short horizon ($H=1$) yields the lowest collision risk (0.62\%) but more aggressive motion (3.000 rad/s), while longer horizons smooth the motion but increase collision risk (up to 4.83\% at $H=100$) due to accumulated mismatch between prediction and rapidly updated observations. Figure~\ref{fig:7dof_dynamic_avoidance} confirms successful avoidance and recovery in the presence of human-induced disturbances.
\begin{table}[t]
\centering
\caption{Quantitative analysis of MPC online control for the 7-DoF manipulator}
\label{tab:7dof_mpc_online}
\begin{tabular}{lcccc}
\toprule
\multirow{2}{*}{Receding Horizon Length} & \multicolumn{3}{c}{Metrics} \\
\cmidrule{2-4}
 & CR (\%)$\downarrow$ &MCI (rad/s)$\downarrow$ & CF(Hz)$\uparrow$ \\
\midrule
$H=1$ & 0.620 & 3.000 & 280.275 \\
$H=10$ & 3.440 & 2.001 & 204.230 \\
$H=50$ & 3.210 & 2.001 & 56.425  \\
$H=100$ & 4.830 & 2.001 & 39.098 \\
\bottomrule
\end{tabular}
\end{table}
\subsubsection{Comparative Analysis}
\begin{table}[t]
\centering
\caption{Comparative trajectory planning performance for the 7-DoF manipulator}
\label{tab:7dof_planning_comparison}
\begin{tabular}{lcccc}
\toprule
\multirow{2}{*}{Algorithms} & \multicolumn{3}{c}{Metrics} \\   
\cmidrule{2-4}
 & CR(\%)$\downarrow$ & TL$\downarrow$ & AT (ms)$\downarrow$ \\
\midrule
RRT-Connect\cite{RRT-connect} & 0.274 & 34.279 & 8.846  \\
BiEST\cite{BiEST} & 0.256 &  51.036 & \textbf{7.715}\\
LazyPRM\cite{LazyPRM} & 0.258 &  8.903 & 422.712\\
KPIECE\cite{KPIECE} & 0.558 &89.549 &42.256\\
CHOMP\cite{CHOMP} & \textbf{0.236} & 31.492 & 641.569 \\
STOMP\cite{STOMP}& 0.312 & 26.792& 796.784\\
TrajOpt\cite{Schulman2014MotionPW}& 0.296 & 23.594 & 191.232 \\
RRT*\cite{RRT*} & 0.342&   5.858 & 501.234 \\
EST\cite{BiEST} & 0.997 &  29.488 & 8.1041 \\
PDST\cite{PDST} & 1.278 &  18.677 & 40.621 \\
InformedRRT*\cite{informed_RRT} & 0.242 &  \textbf{5.754} & 501.454 \\
RRT\cite{RRT} & 1.233 & 19.510& 7.903\\
Proposed Method & 0.561 & 16.758 & 161.117 \\
\bottomrule
\end{tabular}
\end{table}

Table~\ref{tab:7dof_planning_comparison} benchmarks our method against representative planners. The proposed method achieves a favorable balance between trajectory length (16.758 rad) and runtime (161.117 ms), while maintaining low collision risk. Compared with optimization baselines such as CHOMP and STOMP, it produces shorter trajectories with substantially lower computation time, demonstrating the benefit of fast, differentiable C-space distance/gradient queries.
\section{Conclusions}
\label{conclusion}
Experiments show that CSSDF-Net enables effective safety-constrained planning and control with real-time distance/gradient queries. In the 2-DoF planar setting, safe trajectory optimization achieves $0\%$ collision risk (AT $=130.48$ ms), and for the 7-DoF manipulator GPU distance+gradient inference remains fast even at large query scales (e.g., $4.302$ ms at $10^4$ points). These results indicate that the learned C-space distance field provides stable gradients that are directly usable in trajectory optimization and MPC.

Beyond performance, this work advocates a unified and differentiable notion of safety modeled directly in configuration space, covering both self-collision and environment collision within a single geometric representation. By avoiding workspace-to-C-space projections and leveraging a spatial-hashing-based data pipeline, CSSDF-Net targets deployment in unseen scenes without environment-specific retraining, making it a practical backend for vision-guided motion planning in unstructured and dynamic environments.

For future work. Although online updates are supported through repeated point-cloud queries, the model is not explicitly spatiotemporal; extending the formulation toward 4D (space--time) representations is a natural direction for richer dynamic-obstacle modeling. In addition, robustness to sensing noise, partial observability, and calibration errors should be strengthened via uncertainty-aware training and more conservative safety margins. Finally, scaling to dense scenes and multi-agent settings will require more efficient point selection/aggregation and principled coordination mechanisms that couple multiple robots under shared safety constraints.

\bibliographystyle{IEEEtran}
\bibliography{IEEEabrv,reference}

\end{document}